# Structural Design Recommendations in the Early Design Phase using Machine Learning

Spyridon Ampanavos[1]*, Mehdi Nourbakhsh[2], Chin-Yi Cheng[2]

[1] Harvard Graduate School of Design, Cambridge MA 02138, USA
[2] Autodesk Research, San Francisco CA 94105, USA
`sampanavos@gsd.harvard.edu, mehdi.nourbakhsh@autodesk.com,`
`chin-yi.cheng@autodesk.com`

**Abstract.** Structural engineering knowledge can be of significant importance to the architectural design team during the early design phase. However, architects and engineers do not typically work together during the conceptual phase; in fact, structural engineers are often called late into the process. As a result, updates in the design are more difficult and time-consuming to complete. At the same time, there is a lost opportunity for better design exploration guided by structural feedback. In general, the earlier in the design process the iteration happens, the greater the benefits in cost efficiency and informed design exploration, which can lead to higher quality creative results.

In order to facilitate an informed exploration in the early design stage, we suggest the automation of fundamental structural engineering tasks and introduce ApproxiFramer, a Machine Learning-based system for the automatic generation of structural layouts from building plan sketches in real-time. The system aims to assist architects by presenting them with feasible structural solutions during the conceptual phase so that they proceed with their design with adequate knowledge of its structural implications.

In this paper, we describe the system and evaluate the performance of a proof-of-concept implementation in the domain of orthogonal, metal, rigid structures. We trained a Convolutional Neural Net to iteratively generate structural design solutions for sketch-level building plans using a synthetic dataset and achieved an average error of 2.2% in the predicted positions of the columns.

## 1  Introduction

Structure is a fundamental element of a building design. When the structural design is developed in parallel and in coordination with the architectural design, it can inform an architect's decisions and lead to a harmonious integration of the two. However, when structure is not taken into account during the early phase of design, reconciling architectural and structural design can be a cause of delays, conflicts between architects and engineers, and undesirable design compromises.

A study investigating the collaboration between architects and structural engineers conducted in New Zealand in 2009 found that among the primary points of friction are the limited understanding of structural engineering from the side of architects and the late involvement of structural engineers in the project [1].On the other hand, it has been repeatedly argued that the cost of design changes increases the later they are introduced in the process [2, 3]. The term 'cost' is not limited to monetary expenses but can be generalized to the ability of a change to impact the design [4].

Parametric modeling and BIM software have been used by practitioners and researchers to address such collaboration conflicts [4], and specialized software has been used in research and educational settings to facilitate and promote a better understanding between architects and structural engineers [5]. While such solutions have significantly benefited the field, they do not specifically address the conceptual stage of the design.

The conceptual stage of design is commonly described as a divergent process. It is characterized by quick iteration, and often happens outside of a CAD environment, in the form of sketching. In order to achieve a smoother integration of architectural and structural design, structural feedback should be easily available during the conceptual stage. Such feedback cannot and does not need to be precise, as the design itself in this phase lacks precision. In contrast, approximate and directional feedback can be useful for improving a design towards a better solution with respect to its structure.

In this paper, we introduce ApproxiFramer, an automated system with the ability to generate structural design recommendations during the conceptual phase of architectural design. The goal of such recommendations, indicating potential/optimal structural solutions, is to inform the architects' design decisions and ultimately reduce conflicts with the structural engineers when they get involved in the project at a later stage.

A tool targeting the conceptual design phase has to respond to two main challenges. First, the feedback needs to be generated in real-time. Second, the tool should be able to directly handle conceptual sketches without requiring the user to translate them into different software. Recent advances in the field of Machine Learning (ML) have demonstrated an increasing ability to handle irregular types of input data, such as images or sketches. In addition, ML methods have been previously used to successfully accelerate structural design tasks [6–9]. ApproxiFramer employs a machine learning model to tackle both the speed and the integration challenges.

In this paper, we develop and evaluate ApproxiFramer by focusing on a specific structural domain, rigid metal structures. We trained a neural net on a synthetic dataset consisting of sketch-level single floor building plans and their corresponding structural layouts. The neural net generated structural layouts in real-time while achieving an average percentage error of 2.21% in the positions of the structural elements of the test set, confirming the potential of the method for early phase design assistance.

We make two contributions in the area of early-phase design decision support. First, we introduce a method for generating approximate structural solutions for architectural sketches in real-time. Second, we report on the results of an experiment and demonstrate that a machine learning-based system can successfully learn to generalize a consistent set of structural principles.

## 2  Related Work

Some previous work seeking to assist architects in designing buildings that better conform to various performance criteria has employed various forms of optimization [10]. Such works that focus on the early design phase typically combine procedural modeling and simulation software, with the parameters of the generative model being tuned through an optimization algorithm [11, 12]. Shea et al. elaborate on how parametric modeling and engineering performance feedback can be used to improve architectural designs [13]. Optimization is not necessarily the end goal of these processes but rather a tool to automatically construct solutions that can guide the architect towards design improvements [14, 15]. Other work has focused on integrating designers' preferences through interactive optimization [16, 17]. More recently, Hamidavi et al. proposed a system that uses multiple types of structural optimization and BIM modeling to improve the collaboration of architects and structural engineers [18]. However, setting up a good procedural model for an optimization process is non-trivial, and an optimization framework to guide this process has been suggested as well [19].

In practice, optimization and the performance simulations that it relies on are often too time-consuming to be employed in the early design phase. The use of surrogate models has been suggested as a way to accelerate simulations [20]. Tseranidis et al. provide an overview of multiple ML algorithms for the approximation of structural engineering calculations [21]. While most surrogate models are trained to only work with specific parametric models and structural topologies, some research has addressed generalizable models that work with multiple topologies of 3d trusses [9].

Other work has used machine learning to directly approximate optimal solutions. Support Vector Machines have been trained to optimally solve individual modules of a space frame [6], Bayesian nets have been used for bi-directional inference with the goal of identifying the most promising areas of a design space with respect to structural performance [22], and neural nets have been used to predict optimal parameters describing the bracing of a metal frame [8].

In contrast to previous research that has targeted design using parametric models of the geometry, we are proposing a method for structural design approximation that directly uses sketch-level plans. The goal is to provide real-time guidance in the early design exploration during the actual sketching before an idea is formalized into a CAD drawing.

## 3  Method

### 3.1  Approach

ApproxiFramer aims to inform the early phase design exploration in a sketch-based environment through the real-time generation of structural designs. Figure 1 describes how ApproxiFramer can be integrated into such an environment. A user-generated sketch first passes through a pre-processing step that converts a noisy and imprecise input to a clean drawing so that it can be used with our predictive system. This kind of processing is common in commercial design graphics software, so this research considers it given, and further technical elaboration is out of scope.

Consequently, we propose the use of a neural net to solve the problem of real-time structural layout predictions from building plans. Inspired by previous work that suggests the decomposition of structural problems into sub-problems that are easier and more generalizable [8, 23], we do not attempt to estimate the complete structure at once. Instead, we use an iterative approach, only locally solving the problem and predicting a partial structure, gradually extending the solution until no more extensions are necessary. We expect that the neural net will have more chances of identifying patterns when focusing on a small area of the given building at each step, even if every observed building is unique. In order to evaluate and further develop the proposed method, we conducted an experiment where the scope of the problem has been limited, as described in the next section.

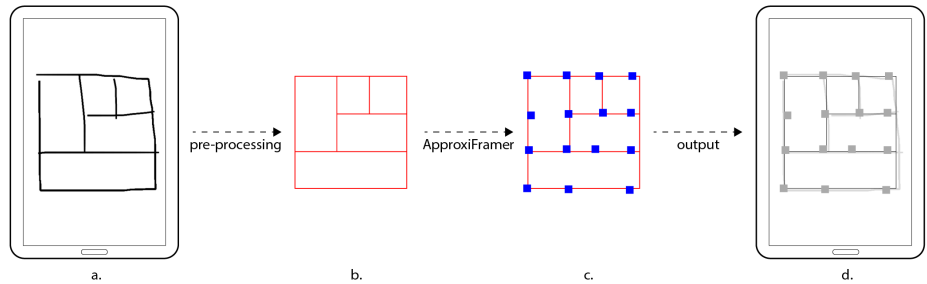

**Fig. 1.** Sketch-based interaction and structural predictions. The user designs a sketch of a plan (a.), the system converts the noisy sketch to a clean drawing (b.) and passes it to ApproxiFramer that predicts the placement of the structural elements (c.). The structural solution is superimposed on the user's initial sketch (d.).

### 3.2  Problem Scope

In general, the design of a structure is informed by a series of specifications and constraints. The type of structural system, materials, available structural members, regulation, and others will all affect the solution of the design, so that the same building design may lead to very different structural designs based on these parameters. The current experiment operates in a constrained space where these parameters are assumed to have fixed values.

In this paper, we focus on rigid metal frames, always connected at right angles. No bracing is typically required for such structures. Selection of the appropriate cross-sections and sizing of the structural elements are outside of the scope of this study and are typically of minor importance during the conceptual design phase. As such, the structures that we design can be easily abstracted to a diagrammatic level. A set of coordinates that indicate the locations of the columns is then a sufficient description of such a frame.

The developed system generates structural layouts for orthogonal, sketch-level, single-floor plans. These plans include exterior and interior walls of the building, both represented by single straight lines that are either horizontal or vertical.

### 3.3 System Architecture

The input of the system is in image format, providing significant flexibility to the user in the selection of design software or medium. In the core of the system lies a convolutional neural net (CNN) that we trained to take an image of a sketch, representing a plan of a building layout, and predict the position of a group of columns. In each iteration, the newly predicted columns are added to the solution, and a new image is rendered that contains both the initial sketch and the columns that have been placed so far. This newly rendered image is then used as the input of the next iteration. Algorithm 1 describes this iterative process.

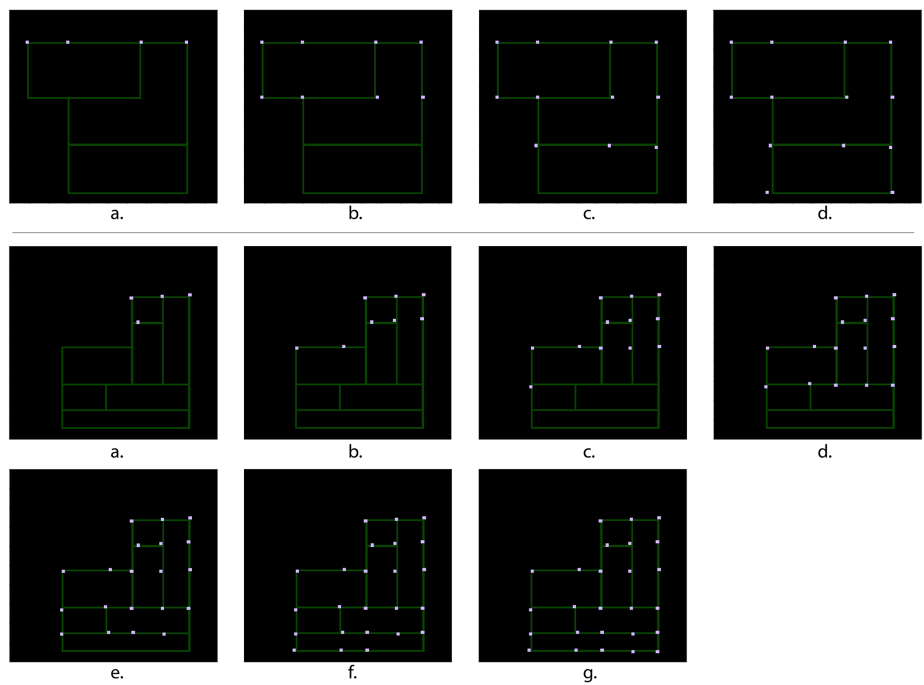

**Fig. 2.** Representative examples of iterative predictions for a smaller structure in four steps (top) and a larger structure in seven steps (bottom).

Each iteration solves a local sub-problem, scanning the building from left to right and from top to bottom, and adding a fixed number[1] of columns to the solution. The columns are assumed in a specific order as well: left to right and top to bottom. The model was trained to output a zero vector when there are no more columns to be placed.

---

[1] The last iteration will add any number of columns between 0 and that fixed number, as needed.

**Algorithm 1.** Predicting positions of all columns.

```
Image ← initialSketch
Columns ← []
repeat
  column ← predictNextColumn(image)
  columns.append(column)
  image ← addColumnToImage(image, column)
until allColumnsHaveBeenPlaced(image, columns)
return image, columns
```

The number of columns for each iteration was defined as n=4, based on initial results after considering alternatives between 1 and 8. Figure 2 demonstrates two examples of structures being predicted in 4 and 7 iterations.

The CNN takes as input an image of 128 X 128 pixels with four channels. Two channels contain the building layout and the already placed columns, and two contain the pixel coordinates, as suggested in [24]. The image passes through three convolutional layers with kernel sizes 7, 3, 3 and strides 2, 2, 2 and a ResNet block [25], followed by two fully connected layers, an LSTM layer [26], and two output layers which are also fully connected. The first output layer (4X2) contains the coordinates of the four predicted columns. The second output layer (4X3) contains the type classification for each of the predicted columns. The possible types are free-standing, column on corner, or column on wall. All layers use ReLu activation functions, except for the ResNet and the output layers. The ResNet uses linear activations and is followed by batch normalization and leaky ReLu. The coordinates output layer uses sigmoid activation since the coordinates are normalized in the range [-1, 1], and the type output layer uses softmax to convert the output to class probabilities. The number of filters is shown in Figure 3, which depicts the structure of the neural net in detail.

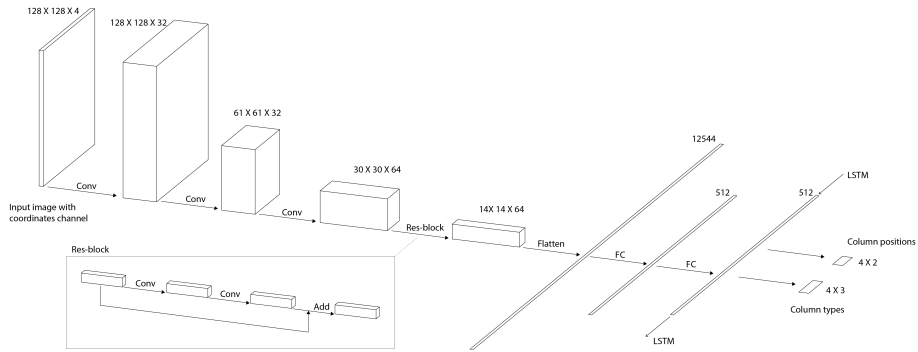

**Fig. 3.** CNN Architecture.

## 3.4 Dataset

An appropriate dataset can be sourced from historical or synthetic data. In general, we expect that given a number of buildings and their corresponding structures that follow a specific set of principles, we can train a CNN to abstract these principles and iteratively generate similar structures for more buildings of the same type. In this work, we generated a synthetic dataset of buildings on an orthogonal grid. The structural layouts were generated using heuristics. While this is not the ideal scenario to demonstrate the power of our system, our focus here is to demonstrate the ability of the system to approximate a set of structural designs, which is expected to generalize to other, more sophisticated datasets as well.

We created 35 building layouts, each building including both exterior and interior walls. This initial set of buildings was augmented through 90-degree rotations, scaling, and translations. For each building, we designed a structural layout using the same heuristics: fitting a grid of columns with a predefined maximum span. The resulting dataset contains 10,000 pairs of buildings and structural layouts. Out of these, we used 9,000 for training and validation, and 1,000 for testing. In order to use the training data with the system's iterative approach, we generated the set of all possible configurations of incremental structural designs for each of the buildings. In the incremental structural designs, we determine the next partial solution - i.e., the next group of columns to be placed - by ordering the columns by x and y. After this process, we ended up with 137,644 training samples and 12,514 testing samples.

## 3.5 Training

In order to obtain a complete structural solution for a design layout, we need to run the model in an iterative way, in each step adding to the observed image the predicted columns of the previous step. However, there are a few challenges in practice. Each time that a column is predicted, the location contains a small error (i.e., the alignment may be one or more pixels off). When this column is added to the input image of the next step, it contributes to a larger error in the next prediction. Eventually, the error accumulates until the model is unable to predict the next column locations in a sensible way.

We used two methods to overcome the problem of the accumulated error. First, we created a new dataset with added noise in the locations of the rendered columns. Adding noise is a technique that has been used with neural nets for data augmentation [27] in order to improve generalization and avoid overfitting. Similarly, by training our model on noisy inputs, we aim to make it robust to inaccuracies during iterative prediction.

Second, we worked towards increasing the output size of each step and, by doing so, reducing the number of iterations that are needed to complete a structural layout. Using an earlier, simpler model, we found that simply increasing the size of the output decreased the performance dramatically when no other major changes were made. However, we were able to get good results when we introduced a residual block and an LSTM layer in the model.

The loss function was defined as the weighted sum of the mean absolute error of the coordinates output layer and the categorical cross-entropy of the column type output layer, with weights 1.0 and 0.2. The network was trained using stochastic gradient descent for 900 epochs.

## 4  Results

On a single run, the CNN model is outputting predictions for four columns. The output includes the column coordinates and the column type (between free-standing, column on corner, or column on wall) (Figure 3). While we found that training using a weighted loss on a combination of the column coordinates and the column types improved the model performance compared to training on column coordinates prediction alone, the column type information is not used during inference, and therefore it is also excluded from the following results.

### 4.1  Single predictions on perfect observations – CNN evaluation

First, we evaluate the model performance on single predictions (i.e., four columns). We use as input all possible partially completed structures from the test set, with a four columns step, and following the ordering by x and y coordinates. The partial completion is done based on the ground truth so that the model is predicting based on a perfect observation.

The model successfully identified when to stop adding new columns 100% of the time. During training, we used a zero vector to indicate the stopping point of the predictions. During inference, we modified the threshold to be a vector where at least one of the x or y coordinates has a value less than 2.

Our CNN achieved an average error of 2.21% in the predicted column positions. This error corresponds to a mean distance of 1.51 pixels between the predicted column locations and the ground truth for our dataset images of 128 X 128 pixels. Figure 4 Left shows the mean distance between predictions and ground truth in relation to the ordering of the columns. We do not observe a significant change in performance as the size of the observed, partially completed structures increases. However, we notice that within each four columns (or column quads) coming from a single prediction, earlier columns tend to have smaller error. This is better captured in Figure 4 Right, where columns have been put in four groups based on their order of appearance within a single prediction. The mean absolute error increases from 1.46 pixels for the first columns of each prediction to 2.14 pixels for the fourth columns. This behavior is attributed to the use of the LSTM layer, which introduces a recurrent architecture in the model. Each column of a single prediction depends on the previously estimated columns of the same prediction, and as a result, the error accumulates.

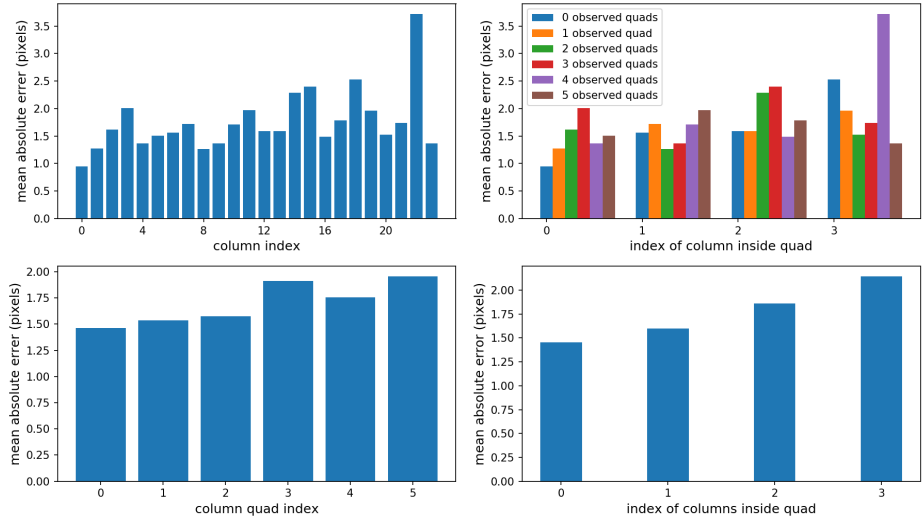

**Fig. 4.** Left Top: Mean error for first 24 columns. Left Bottom: Mean error for first six prediction steps, where one prediction step outputs four columns (or one "column quad"). Predictions on clean observations. Right Top: Mean absolute error of first six quad predictions, grouped by order inside quad – the first group is first columns of 6 quad predictions, the second group is second columns, etc. Right Bottom: Mean absolute error of same order columns. Predictions on clean observations.

### 4.2 Iterative Predictions – System Evaluation

Next, we evaluate the system performance on the goal task, which is to estimate all columns for each building. This is accomplished by using our CNN in an iterative way, where each step relies on the output of the previous prediction.

The system predicted the correct number of columns 95.3% of the time. Out of the 1000 buildings of the test set, 47 were solved with fewer or with more columns than the ground truth. These buildings have been removed from the report of the rest of the metrics described below.

The system achieved an average error of 2.21% in the predicted column positions. The mean distance of the predictions from the ground truth among all iterations was 1.84 pixels. Figure 5 shows in green the mean distance of the predictions from the ground truth in relation to the ordering of the columns. We observe that the error increases for inputs with more columns in the already completed structure. In the previous subsection 4.1, we found that the size of the completed structure only has a minimal effect on the CNN performance. Therefore, we attribute this error increase to the accumulated error from the previously predicted columns that are used in the input of the new predictions.

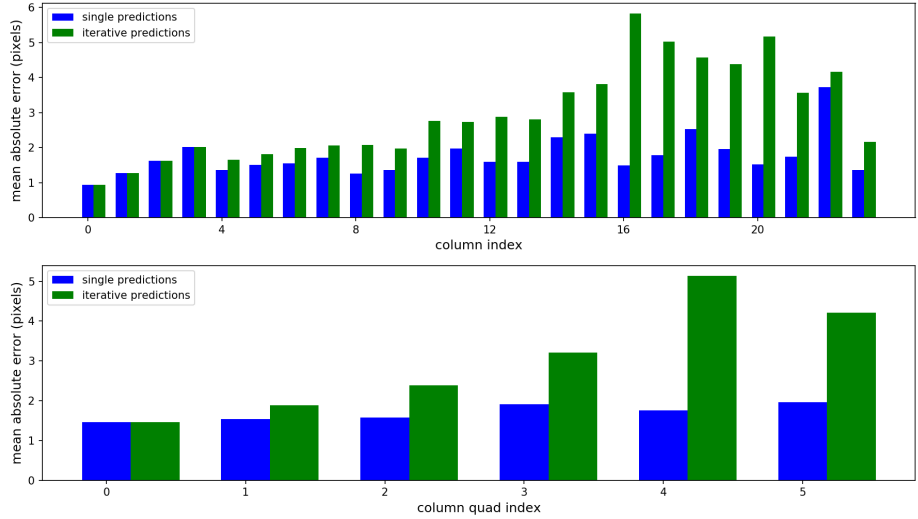

**Fig. 5.** Top: Mean absolute error for first 24 columns. Bottom: Mean absolute error for the first six prediction quads.

## 5 Discussion

### 5.1 ML as an Approximation Means for Early Phase Structural Assistance

Our model performs very well on single predictions, maintaining a very low average error (2.21%) and producing results that are visually coherent. The system also maintains a low average error on iterative predictions (2.21%). In practice, it predicts all column positions well on a large subset of the test set and only fails to output reasonable results while iterating on some building designs. This happens as the model remains susceptible to noisy observations of previous predictions. Currently, the performance tends to decrease both with repeated iterations (Figure 4 Left and Figure 5), as well as with later outputs of a single run (Figure 4 Right). Further investigation is needed on the potential of the two approaches and the optimal combination of them.

The results suggest that the proposed method, which relies on machine learning techniques, constitutes a promising approach for the automatic suggestion of structural designs for early phase architectural sketches or drawings. Once a trained model is loaded, our system only needs a few milliseconds to generate such a structure. Therefore, we believe that a system like ApproxiFramer can provide valuable design assistance during the conceptual design phase.

Furthermore, ApproxiFramer could be combined with other ML methods that propose optimal cross sections of columns and beams [28], based on structural skeletons similar to the ones that our system generates. It can also be used as a complementary tool to parametrization and optimization methods such as the one introduced in SketchOpt [29], providing early, quick estimates before a structural optimization is run.

### 5.2   Iterative approach vs. end-to-end model

The ApproxiFramer system relies on the iterative use of a neural net to predict a complete structure. Early experimentation results, as well as the increasing error between the first and last predictions of a single model inference (Figure 4 Right Bottom), suggest that the current model architecture is not appropriate for the end-to-end prediction of complete structures.

Apart from performance considerations, we believe that the iterative approach has other advantages, too, over a whole-structure prediction. The model's ability to complete partial structures could be potentially leveraged to interactively guide the design of structure as well as space, following the design paradigm of interactive optimization [17]. Combined with a different sub-problem parsing strategy in the future, e.g., one where subsequent iterations have increased level-of-detail, this would allow the designer to lead the system towards a specific direction, for example, by modifying the outputs of the initial steps of the structure prediction.

### 5.3   Generalizability

In this paper, we demonstrated the feasibility of the suggested method in the domain of rigid metal frames connected at right angles. Functioning inside this domain, we were able to simplify a structural design to a set of columns, assuming that beams can be added in a post-processing step using simple heuristics. Even though we considered single-floor plans, the method is easily generalizable to low-rise buildings with a typical plan repeated in all floors above the ground. We expect that our method is also generalizable to different structural systems. However, appropriate modifications will have to be made to accommodate the potential use of multiple types of structural elements in more complex domains. For instance, we have already demonstrated the prediction of labels associated with each column, and it is not difficult to imagine how such labels could be used as classes of multiple types of structural elements.

The dataset used in this work contains orthogonal designs with heuristically generated structures; however, we expect the proposed method to be generalizable to different datasets and human-generated or computationally optimized structures.

### 5.4   Limitations

The results indicate that the column positioning tends to be noisy, something that may be easier noticeable for columns that should be placed on wall intersections. This is not necessarily an issue in the specific context of early phase sketching since the user's sketches are expected to be similarly rough, and precision is not the goal at this stage. Nevertheless, a post-processing step might be able to fine-tune the positions at a local level.

Last, larger structures are currently more difficult to solve, mainly because of the accumulated error. A larger dataset and different data augmentation techniques are expected to improve this performance.

## 6      Conclusion

We introduced a method for early phase design assistance with respect to structure, with the goal of promoting more informed design decisions early on and better preparing architects for later stage collaboration with structural engineers. We described a system that can predict structural layouts of single-story building designs from diagrammatic sketches and trained a CNN that performs this task in an iterative way. While we achieved satisfactory performance on our test set, our model only consists a first step to applying this approach to the real world. Further improvements are required in terms of robustness and the ability to take into account structural constraints and parameters.

Future work may explore how to interactively address architectural aspects of the structure by adding the designer in the loop between prediction iterations. Alternative ways of parsing the overall problem into smaller sub-problems may also be investigated in relation to different structural systems. The addition of a fine-tuning step can be investigated as a way to reduce the impact of small inaccuracies in the positions of previously placed columns. Finally, the use of a synthetic dataset generated through optimization or a dataset from historical data - i.e., from a structural engineering practice - will be a significant step towards deploying such a system in the real world.

**Acknowledgements**. We would like to express our gratitude to Mohammad Keshavarzi for his help with the synthetic data preparation process.